\documentclass[conference,a4paper]{APSIPA2025}
\usepackage{graphicx}
\usepackage{multirow}
\usepackage{threeparttable}
\usepackage[backend=biber]{biblatex}
\usepackage{amsmath,amssymb,amsfonts}
\usepackage{amsmath}
\usepackage{algorithmic}
\usepackage{textcomp}
\usepackage{bbm}
\usepackage{booktabs}
\usepackage{colortbl}
\def\BibTeX{{\rm B\kern-.05em{\sc i\kern-.025em b}\kern-.08em
    T\kern-.1667em\lower.7ex\hbox{E}\kern-.125emX}}
\definecolor{softblue}{HTML}{CCE5FF} 
\definecolor{softgreen}{HTML}{CBFFCC}

\usepackage{geometry}
\usepackage{fancyhdr}

\fancypagestyle{firststyle}{
  \fancyhf{}
  \fancyhead[C]{2025 Asia Pacific Signal and Information Processing Association Annual Summit and Conference (APSIPA ASC)}
}

\begin{document}

\title{Identifying Speaker Information in Feed-Forward Layers of Self-Supervised Speech Transformers}

\author{
\authorblockN{
Tzu-Quan Lin\authorrefmark{1} and
Hsi-Chun Cheng\authorrefmark{1} and 
Hung-yi Lee\authorrefmark{1} and 
Hao Tang\authorrefmark{2}
}

\authorblockA{
\authorrefmark{1} Graduate Institute of Communication Engineering, National Taiwan University, Taiwan \\
E-mail: tzuquanlin@gmail.com, leo5470@gmail.com, tlkagkb93901106@gmail.com}

\authorblockA{
\authorrefmark{2} Centre of Speech Technology Research, University of Edinburgh, United Kingdom \\
E-mail: hao.tang@ed.ac.uk}
}

\maketitle
\thispagestyle{firststyle}
\pagestyle{fancy}

\begin{abstract}
In recent years, the impact of self-supervised speech Transformers has extended to speaker-related applications.  
However, little research has explored how these models encode speaker information.  
In this work, we address this gap by identifying neurons in the feed-forward layers that are correlated with speaker information.  
Specifically, we analyze neurons associated with k-means clusters of self-supervised features and i-vectors. 
Our analysis reveals that these clusters correspond to broad phonetic and gender classes, making them suitable for identifying neurons that represent speakers.
By protecting these neurons during pruning, we can significantly preserve performance on speaker-related task, demonstrating their crucial role in encoding speaker information.  
\end{abstract}

\section{Introduction}

Self-supervised Transformers have enjoyed a great success in learning
speech representations that can be applied to a wide range of tasks~\cite{chung2019unsupervised, ling2020decoar, liu2021tera, hsu2021hubert, baevski2022data2vec, chen2022wavlm, liu2024dinosr}.
Their main application has largely been on automatic speech recognition,
but the impact of self-supervised Transformers has extended
to speaker-related tasks~\cite{fan2021exploring, wang2022finetuned, chen2022largescale},
rivaling the state of the art~\cite{miara2024supervised}.
Despite the successes, few (except~\cite{chowdhury2024what}
and~\cite{mohamed2024orthogonality}) have analyzed how speaker
information is encoded in these models.
Even in~\cite{chowdhury2024what} and~\cite{mohamed2024orthogonality},
the analyses have been limited to probing classifiers.
There is generally a lack of studies and tools for analyzing speaker information
in self-supervised speech Transformers.

A recent study~\cite{lin2022compressing} has shown that compressing the
feed-forward layers (also known as FFNs) in Transformers is particularly detrimental
for speaker tasks, hinting that speaker information is more prominent
at feed-forward layers compared to other parts of the Transformer.
The finding provides us an approach to studying speaker information through pruning.
In addition, the interpretation of feed-forward layers
as key-value memory~\cite{gema2020transformer} has proven useful for analyzing 
the feed-forward layers in Transformers.
In particular, given a property of interest (such as phones, gender, or pitch),
one can identify a set of neurons in FFNs that is reponsible for representing
a property~\cite{lin2024property}.
The combination of pruning and identifying property neurons
provides a novel approach to analyzing feed-forward layers in Transformers.

In this work, we study speaker information in self-supervised
Transformers by identifying property neurons related to speakers.
We take the aforementioned pruning methodology as our evaluation:
a set of neurons are deemed important if pruning them leads to severe degradation
in performance.
In other words, the performance should be largely preserved
if the set of important neurons are \emph{protected} during pruning.
If we are able to identify a set of neurons that does not hurt
a speaker task much when protected during pruning, then that
set of neurons are the ones where the speaker information is
prominent.

There are a few challenges when adopting this methodology.
The first challenge is that as the number of classes within a property increases,
the number of identified neurons decreases.
As we will see later in the experiments,
this seems to be an inherent drawback of the approach proposed in~\cite{lin2024property}.
Given the large number of speakers in a typical speech dataset,
using speaker IDs directly would result in very few property neurons being identified.
A subsequent challenge is to decide what alternative property to use for identifying speaker-related neurons,
especially under the constraint that speaker labels are unavailable.

To address this, we opt for the classes discovered by k-means (with a small $k$)
on self-supervised features and i-vectors~\cite{dehak2010front}.
While x-vectors~\cite{snyder2018x} offer stronger performance, they require supervised training.
In contrast, i-vectors are label-free, aligning better with our unsupervised pipeline.
Inspired by prior works~\cite{liu2018speaker, liu2019introducing} showing that phonetic information can aid the learning of speaker embeddings,
we hypothesize that identifying speaker-related neurons requires both phonetic and speaker information.
Accordingly, we use both SSL and i-vector-based clusters, which we expect to correspond to
broad phonetic and speaker classes, respectively.
This unsupervised design allows us to identify speaker-relevant neurons without relying on any labels.
An added benefit of this approach is that no labels are needed
when identifying neurons with the k-means clusters.

Our experimental results show that by protecting the neurons discovered
using clusters on self-supervised features and i-vectors, we can significantly
preserve speaker information when pruning FFNs and improve pruning effectiveness.
Our approach, without requiring labels, rivals the approach proposed
in~\cite{lin2024property} which does use phone and gender labels.
Our approach works especially well in the one-shot pruning setting.
We are able to prune about 70\% of the total parameters in FFN in a
single step without causing significant performance degradation.

\begin{figure*}
  \centering
  \begin{center}
  \includegraphics[width=\linewidth]{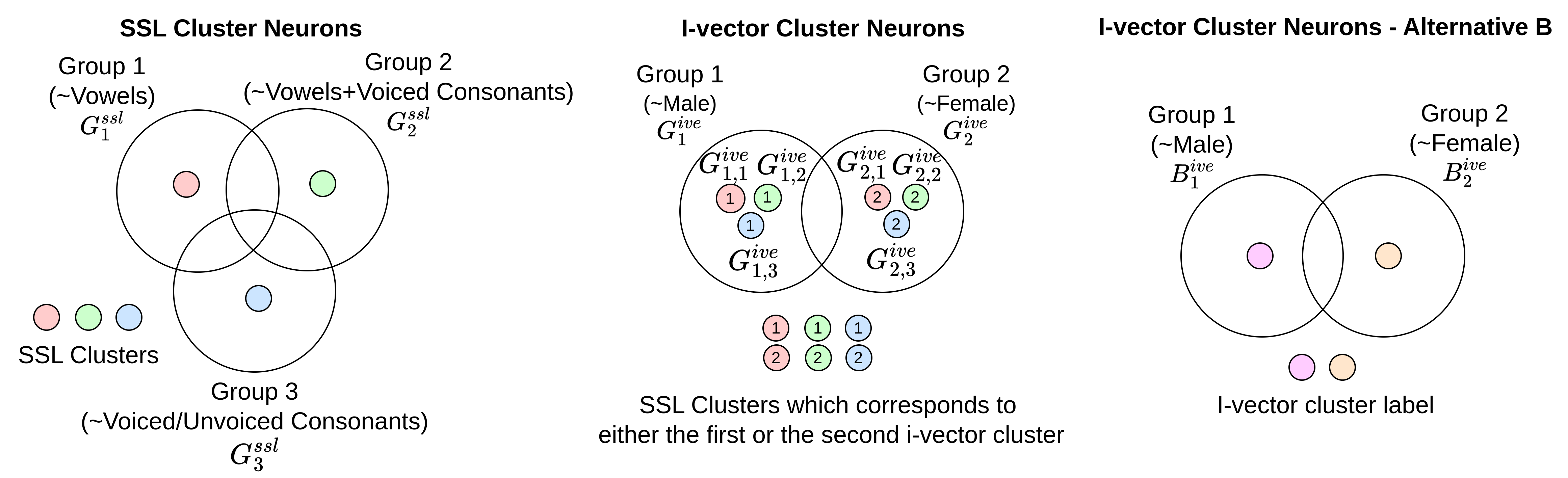}
  \end{center}
  \vspace{-1.5em}
  \caption{An example showing the set relationships of SSL cluster neurons (left),
    i-vector cluster neurons (middle), and one of the possible
    alternatives for identifying i-vector cluster neurons (right).}
  \label{fig:SSL-Neurons}
\end{figure*}

\section{Methodology}
\label{sec:methodology}

In thie section, we review the approach proposed in~\cite{lin2024property}
and discuss the extension made in this work.
At the high level, to find a set of neurons, we need to decide what it means
for a neuron to be activated and how it correlates with a property.

Whether a neuron is activated can be determined by looking at individual frames.
For a frame $v \in \mathbb{R}^d$, a neuron is considered activated when
its value is among the top $\lambda$\% in the frame.
Formally, a frame is binarized into
\begin{align}
b(v) = \begin{bmatrix}
  \mathbbm{1}_{v_1 > \text{top $\lambda$\%} (v)} &
  \cdots &
  \mathbbm{1}_{v_D > \text{top $\lambda$\%} (v)}  
\end{bmatrix}^\top,
\end{align}
where $\lambda$ is a hyperparameter.
This binary vector is called the activation pattern in~\cite{lin2024property}.
Once the activation is determined, we can correlate the activations
with the properties of interest.

\subsection{SSL Cluster Neurons}
\label{sec:SSL-Cluster-Neurons}

As argued in the introduction, we run k-means on the self-sueprvised features
to obtain clusters and correlate the activation patterns with the clusters.
Spefically, for a cluster $c$, we define $G^{\text{ssl}}_{c}$ to be the set of neurons
\begin{align}
\Big\{ d: p([b(v)]_d = 1 \mid \text{frame $v$ is in cluster $c$}) > \rho\% \Big\},
\end{align}
where $\rho$ is a thresholding hyperparameter.
The probability is computed with simple counts of co-occurrences.
The classes used in~\cite{lin2024property} are phones and the discovered neurons
form clusters of vowels, consonants, and semi-vowels.
We will later in the experiments show that clusters of self-supervised features
discovered by k-means already form a good separation of these broad categories.

Similar to the findings in~\cite{lin2024property}, we find that certain neurons
are correlated with speech, and get activated with all the clusters.
We follow~\cite{lin2024property}, selecting the non-overlapping
set among different clusters
\begin{align}
P_{\text{ssl}} = \bigcup_{c=1}^{k_{\text{ssl}}} G_c^{\text{ssl}} - \bigcup_{1 \leq c < j \leq k_{\text{ssl}}} G_c^{\text{ssl}} \cap G_j^{\text{ssl}},
\label{eq:p_ssl}
\end{align}
where $k_{\text{ssl}}$ is the number of SSL clusters.
We refer to the set of neurons $P_{\text{ssl}}$ as SSL cluster neurons.

\subsection{i-vector Cluster Neurons}
\label{sec:ivector-neurons}

The approach of how neurons are discovered is general and can be
easily extended by conditioning them on other properties of speech.
Since our focus is on analyzing the speaker information,
we further cluster the i-vectors of utterances and label all the frames
in an utterance with their i-vector cluster ID.
This leads to further conditioning of the neurons,
and we define the set of neurons $G^{\text{ive}}_{g, c}$
for an i-vector cluster $g$ and SSL cluster $c$ as
\begin{align}
\Big\{ d : p( [b(v)]_d = 1 \mid {}& \text{frame $v$ is labeled i-vector cluster $g$,} \notag\\
  & \quad \text{and SSL cluster $c$}) > \rho\% \Big\}.
\end{align}
To obtain the neurons conditioned just on i-vectors,
the approach taken by~\cite{lin2024property} is taking the intersection
of all SSL clusters
\begin{align}
G^{\text{ive}}_g = \bigcap_{c = 1}^{k_{\text{ssl}}} G^{\text{ive}}_{g, c}.
\label{eq:ivector-group-neurons}
\end{align}
Given the formal definitions, we can now see that as $k_{\text{ssl}}$ increases,
the discovered neurons naturally decreases.

It is not entirely clear why the neurons need to be conditioned on SSL clusters and then later taking the intersection.
There are a few alternatives that can achieve a similar need.

\vspace{0.5em}\noindent\textbf{Alternative A}\hspace{0.25em}
We can take the union
\begin{align}
A^{\text{ive}}_g = \bigcup_{c = 1}^{k_{\text{ssl}}} G^{\text{ive}}_{g, c}. 
\end{align}
instead of intersection in Eq.\ref{eq:ivector-group-neurons}.

\vspace{0.5em}\noindent\textbf{Alternative B}\hspace{0.25em}
We can drop the conditioning on SSL clusters altogether
and define
\begin{align}
B^{\text{ive}}_g = \Big\{ d : p ( [b(v)]_d = 1 \mid {}& \text{frame $v$ is labeled} \\
    & \quad \text{i-vector cluster $g$}) > \rho\% \Big\}. \notag
\end{align}

\vspace{0.5em}\noindent\textbf{Alternative C}\hspace{0.25em}
We can follow Alternative B but further intersect SSL cluster neurons
outside the probability
\begin{align}
C^{\text{ive}}_g = B^{\text{ive}}_g \cap \bigcap_{c=1}^{k_{\text{ssl}}}G^{\text{ssl}}_{c}.
\end{align}

All of the above alternatives provide a set of neurons related to i-vector clusters.
Regardless of how the neurons related to i-vectors are discovered,
the last step is still to find the non-overlappting ones
\begin{align}
P_{\text{ive}} = \bigcup_{g=1}^{k_{\text{ive}}} G_g^{\text{ive}} - \bigcup_{1 \leq g < j \leq k_{\text{ive}}} G_g^{\text{ive}} \cap G_j^{\text{ive}},
\label{eq:p_ive}
\end{align}
where $k_{\text{ive}}$ is the number of i-vector clusters.
Here, $G^{ive}_g$ can be replaced with any of the three alternatives
mentioned above.
It is not immediately clear why we would choose one or the other,
and we will look into these alternatives in the experiments.
Figure \ref{fig:SSL-Neurons} illustrates how the neurons are identified
and their set relations.

\subsection{Protecting Neurons while Pruning}

There are two feed-forward layers in a Transformer block.
We prune the 3072 hidden dimensions after the first feed-forward layer
by removing rows from the first layer and the corresponding columns in the second layer.
Following~\cite{han2015learning}, we prune the dimensions (i.e., neurons) that have
small $\ell_1$-norm of the weights.
Given the set of neurons we discovered in previous sections,
we can decide to keep (or protect) them even when they should have been pruned.

We choose two pruning approaches as our evaluation: one-shot pruning~\cite{lee2018snip} and iterative pruning~\cite{han2015learning, frankle2018lottery}.
For both pruning approaches, we have a target sparsity to achieve.
One-shot pruning prunes the neurons that have low $\ell_1$ of the weights
to the target sparsity in one step.
Iterative pruning, however, iterates between pruning and fine-tuneing,
using a schedule to gradually prune until the sparsity reaches the target.

\section{Experimental Setting}

In the experiments, we will use the methodology outlined in the introduction,
identifying neurons that are potentially important for encoding speakers 
and later protecting them during pruning.
We use the SUPERB benchmark~\cite{yang2021superb},
specifically speaker identification, to probe
the speaker information of self-sueprvised speech Transformers. 
In addition to speaker identification (SID), we will also look at
phoneme recognition (PR), emotion recognition (ER), and intent classification (IC).
The self-supervised models we study are MelHuBERT~\cite{lin2023melhubert}
and wav2vec 2.0~\cite{baevski2020wav2vec}, where MelHuBERT takes mel spectrograms
as input and wav2vec 2.0 takes raw waveform samples.
Our method is also applicable to other models such as HuBERT~\cite{hsu2021hubert} and WavLM~\cite{chen2022wavlm}, but their pre-training targets are not publicly released, making continued pre-training and pruning difficult to evaluate.

For identifying the neurons,
we set $\lambda=1\%$ as the threshold to compute the activation patterns
and $\rho=1\%$ for correlating activation patterns with clusters.
We run k-means on the 9th layer of MelHuBERT to obtain the SSL clusters. 
Recall that we hypothesize in the introduction that speaker-related neurons may rely on both phonetic and speaker information.  
We therefore intentionally select the 9th layer, as it produces phonetically prominent clusters.
We use the same frame-wise cluster assignments for both MelHuBERT and wav2vec 2.0 features.

For iterative pruning, we prune 128 dimensions every 25,000 steps
and continue pretraining the model after each pruning step using a learning rate of 1e-5. 
We use a batch size of 4 for MelHuBERT and 12 for wav2vec 2.0.
Pruning continues until 512 hidden dimensions remain in the FFNs.
For one-shot pruning, we first identify the SSL cluster neurons
and i-vector cluster neurons in the unpruned model,
then prune all other neurons in a single step. 
Since the number of neurons identified vary across layers,
different layers may retain different numbers of dimensions after one-shot pruning. 
To ensure a fair comparison, we prune the model to the same average number
of dimensions across pruning methods.
Specifically, the model is pruned to an average of 867 dimensions per layer
for MelHuBERT and 780 dimensions per layer for wav2vec 2.0.

\begin{table*}
\caption{
The results for iterative and one-shot pruning on MelHuBERT and wav2vec 2.0.
We use ``$\setminus$ cluster neurons'' to denote pruning
while protecting the cluster neurons in this work,
and ``$\setminus$ property neurons'' to denote pruning while
protecting property neurons~\cite{lin2024property}.
Iterative and one-shot pruning results are not directly comparable
due to differences in model sparsity.
Note that Lin \text{et al.}~\cite{lin2024property} use labels for
identifying property neurons, while ours is purely unsupervised.
}
\centering
\scalebox{0.92}{
\renewcommand{\arraystretch}{1.5}
\begin{tabular}{c|l|cc|cc|cc|cc}
 \hline
  \multirow{2}{*}{Model} & \multirow{2}{*}{Pruning Method} & \multicolumn{2}{c|}{SID (ACC\%)$\uparrow$} & \multicolumn{2}{c|}{PR (PER\%)$\downarrow$} & \multicolumn{2}{c|}{ER (ACC\%)$\uparrow$} & \multicolumn{2}{c}{IC (ACC\%)$\uparrow$} \\ \cline{3-10} 
 & & Iterative & One-Shot & Iterative & One-Shot & Iterative & One-Shot & Iterative & One-Shot \\ \hline
 \multirow{4}{*}{MelHuBERT} 
  & unpruned & \multicolumn{2}{c|}{63.96} & \multicolumn{2}{c|}{8.17} & \multicolumn{2}{c|}{57.52} & \multicolumn{2}{c}{72.24} \\ \cline{2-10}
 & regular pruning & 51.04 & 54.08 & 12.28 & 9.11 & 54.96 & 55.92 & 59.32 & 67.65 \\
 & $\setminus$ cluster neurons & 52.66 & \textbf{62.96} & 12.22 & 9.11 & \textbf{57.52} & \textbf{58.67} & 68.81 & \textbf{82.64} \\ \cline{2-10}
& $\setminus$ property neurons~\cite{lin2024property} & \textbf{54.10} & 58.27 & \textbf{10.8} & \textbf{9.00}& 57.11 & 54.82 & \textbf{71.13} & 68.78 \\ \hline
 \multirow{4}{*}{wav2vec 2.0} 
& unpruned & \multicolumn{2}{c|}{75.18} & \multicolumn{2}{c|}{5.74} & \multicolumn{2}{c|}{63.43} & \multicolumn{2}{c}{92.35} \\ \cline{2-10}
 & regular pruning & 57.41 & 63.42 & 14.54 & 11.17 & 59.34 & 60.77 & 89.45 & 91.03 \\
 & $\setminus$ cluster neurons & \textbf{58.34} & \textbf{70.67} & 14.34 & 11.26 & 59.43 & \textbf{62.89} & \textbf{94.06} & \textbf{95.65} \\ \cline{2-10}
 & $\setminus$ property neurons~\cite{lin2024property} & 57.35 & 66.05 & \textbf{14.11} & \textbf{9.70} & \textbf{61.39} & 61.28 & 91.22 & 91.88 \\ \hline
\end{tabular}
}
\label{tab:main-result}
\vspace{-1em}
\end{table*}

\section{Results}
\label{sec:results}

Below we will present three sets of results.
We will first study the k-means clusters and how they capture
broad phone classes.
Next, we will present the comparison between protecting the neurons
we discovered and regular pruning.
Additionally, we also compare to the approach in~\cite{lin2024property},
as our approach is largely derived from theirs.
Finally, we will study the impact of the hyperparameters, in particular,
how the number of clusters and the different alternatives of discovering
neurons impact pruning.

\subsection{Interpreting SSL and i-vector Clusters}

\begin{figure}
  \centering
  \begin{center}
  \includegraphics[width=\linewidth]{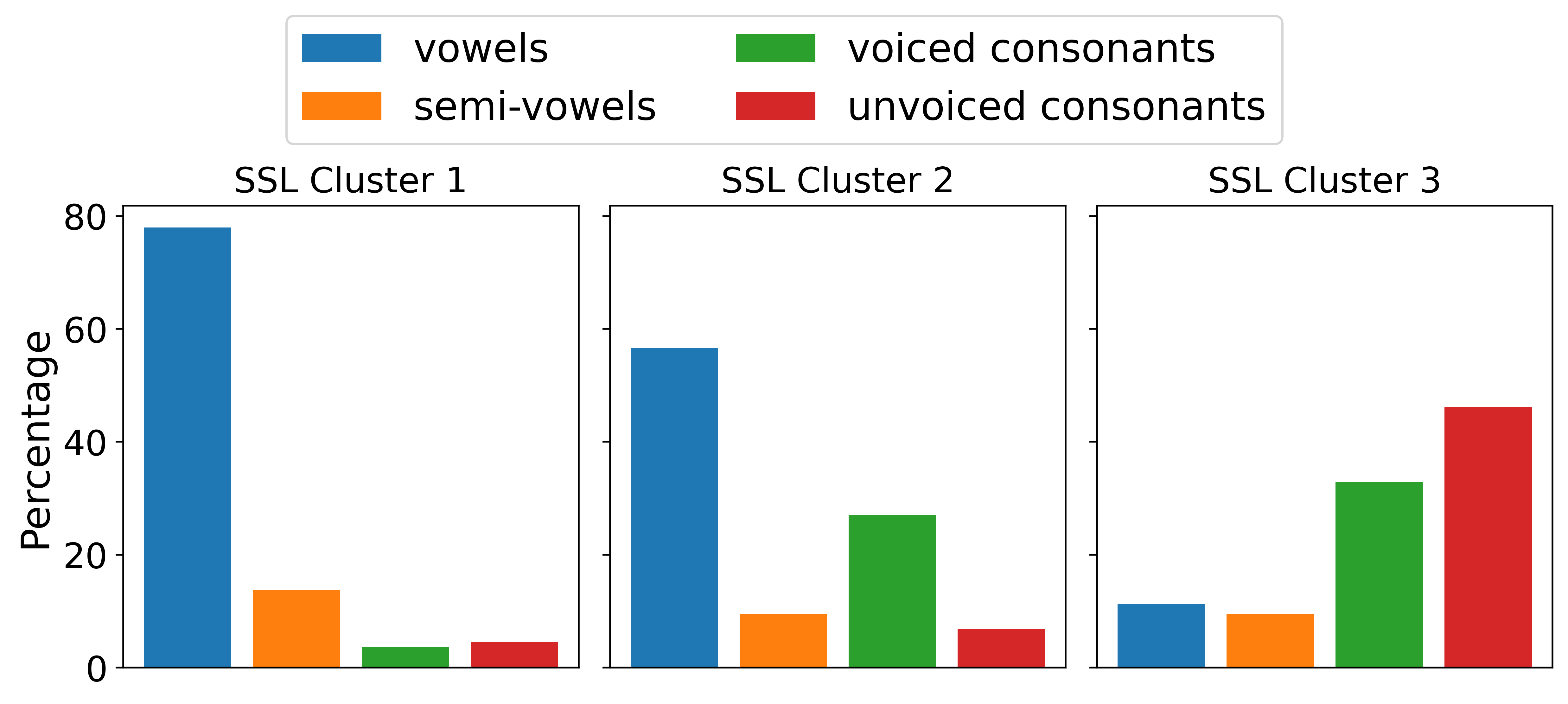}
  \end{center}
  \vspace{-1.5em}
  \caption{Amounts of broad phone classes within the clusters
    produced by running k-means on self-supervised features with $k_{\text{ssl}}=3$.
    \label{fig:ssl-clusters}}
  \vspace{-1em}
\end{figure}

For the SSL clusters, given their rich phonetic information~\cite{yang2021superb, wells2022phonetic, martin2023probing, choi2024self} and their main applications in automatic speech recognition~\cite{baevski2020wav2vec, hsu2021hubert, baevski2020effectiveness, gao2022self}, we hypothesize the clusters are likely to be correlated with broad phone classes.
Figure~\ref{fig:ssl-clusters} shows the result of running k-means on MelHuBERT features
with $k_{\text{ssl}}=3$.
We find that the first cluster predominantly corresponds to vowels,
the second includes vowels and voiced consonants,
and the third mainly contains voiced and unvoiced consonants.
This result is very much aligned with the clustering results shown
in~\cite{lin2024property}, where vowels and consonants are separated
with semi-vowels inbetween the two.
For the i-vector clusters, when $k_{\text{ive}}=2$, they closely match the gender classes.
The first cluster has a 97.23\% purity of being male, while the other cluster
has a purity of 89.82\% of being female.
This again aligns well with the gender results in~\cite{lin2024property}. 

The decision to use a small number of clusters has been discussed in previous sections.
The main issue is that the number of discovered neurons decreases significantly
when using a large number of clusters, as is shown in Eq.~\ref{eq:ivector-group-neurons}.
We will study this problem empirically later, and
this seems to be an inherent problem of the approach in~\cite{lin2024property}.

\subsection{Protecting Neurons during Pruning}

We present our main results in Table~\ref{tab:main-result}.
While all tasks suffer after pruning, speaker identification (SID) suffers the most
by as much as 23.6\% relative for wav2vec 2.0 after iterative pruning.
Protecting both cluster neurons and property neurons help recover
the performance loss due to pruning.
However, protecting the cluster neurons is on par with protecting property neurons,
while being significantly better in one-shot pruning, recovering most of the performance
loss.

Despite not being our goal, protecting cluster neurons also brings certain benefits
for emotion recognition (ER) and intent classification (IC).
In one-shot pruning, our approach can even get achieve a better performance
than the unpruned results.
We suspect that the unpruned results suffer a mild overfitting, and
pruning can serve as a regularization and brings improvement~\cite{bu2020information}.
Protecting cluster neurons has little effect on phone recognition (PR).

Overall, we are able to prune about 70\% of the paramameters with
one-shot pruning while maintaining most of the performance for
speaker identification.

\subsection{Analysis and Ablation Study}
\label{sec:ablation}

To gain further understanding of how the hyperparameters affect the identified
neurons and the pruning performance,
in this section, we conduct analysis and ablation studies for the proposed approach. 
For simplicity and consistency, all experiments here are done on MelHuBERT
with iterative pruning. 

\begin{table*}
\caption{Analysis and ablation study for the proposed approach.
The subscript denotes the differences when changing one of the design decisions
comparing to the proposed approach.}
\centering
\scalebox{1.0}{
\renewcommand{\arraystretch}{1.25}
\begin{tabular}{c|c|c|c|c}
 \hline
      & SID (ACC\%)$\uparrow$ & PR (PER\%)$\downarrow$ & ER (ACC\%)$\uparrow$ & IC (ACC\%)$\uparrow$ \\ \cline{1-5} 
 $k_{\text{ssl}}=100$ & 49.19\textsubscript{-3.47} & 14.20 \textsubscript{+1.98} & 55.36\textsubscript{-2.16} & 60.61\textsubscript{-8.20} \\ \cline{1-5} 
 $k_{\text{ssl}}=39$ & 49.98 \textsubscript{-2.68} & 14.26 \textsubscript{+2.04} & 54.94\textsubscript{-2.58} & 60.79\textsubscript{-8.02} \\ \hline \hline 
 without SSL clusters & 48.84\textsubscript{-3.82} & 14.15\textsubscript{+1.93} & 55.29\textsubscript{-2.23} & 64.62\textsubscript{-4.19} \\ \cline{1-5} 
 without i-vector clusters & 49.86\textsubscript{-2.80} & 13.89\textsubscript{+1.67} & 57.23\textsubscript{-0.29} & 69.62\textsubscript{+0.81} \\ \hline \hline 
 Alternative A & 51.84\textsubscript{-0.82} & 13.78\textsubscript{+1.56} & 56.17\textsubscript{-1.35} & 69.89\textsubscript{+1.08} \\ \cline{1-5} 
 Alternative B & 51.21\textsubscript{-1.45} & 13.48\textsubscript{+1.26} & 56.63\textsubscript{-0.89} & 71.55\textsubscript{+2.74} \\ \cline{1-5} 
 Alternative C & 50.44\textsubscript{-2.22} & 13.79\textsubscript{+1.57} & 55.27\textsubscript{-2.25} & 62.53\textsubscript{-6.28} \\ \hline \hline
 Proposed Method & 52.66 & 12.22 & 57.52 & 68.81 \\ \cline{1-5} 
\end{tabular}
\label{tbl:analysis-ablation}
}
\vspace{-1.2em}
\end{table*}

\begin{figure}
  \centering
  \begin{center}
      \includegraphics[width=\linewidth]{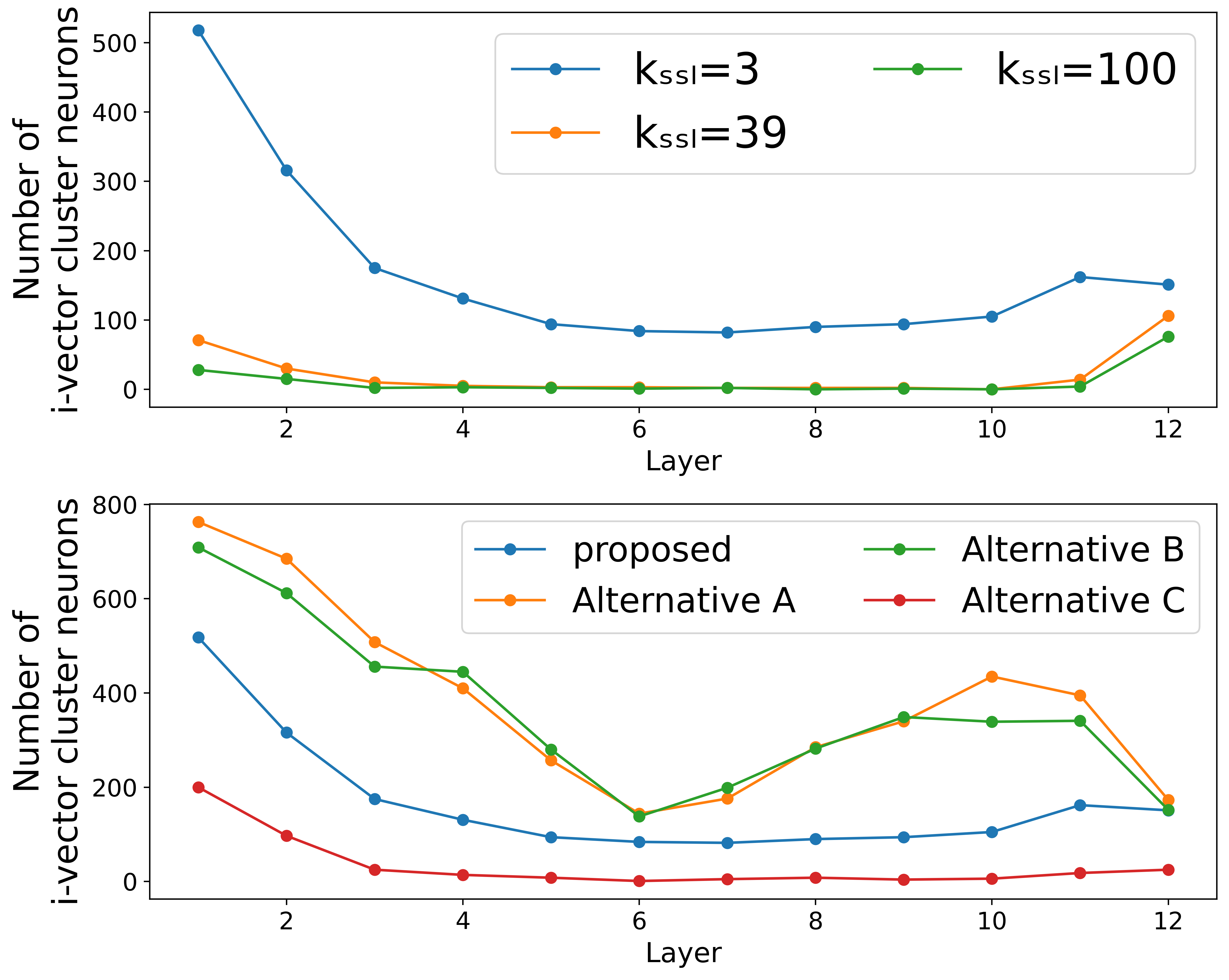}
  \end{center}
  \vspace{-1.5em}
  \caption{The number of i-vector cluster neurons in different layers for different number of SSL clusters $k_{\text{ssl}}$ (top) and different alternatives to identify i-vector cluster neurons (bottom).
   \label{fig:num-ivector-neurons}}
  \vspace{-1.2em}
\end{figure}

\vspace{0.5em}
\noindent\textbf{Effect of Different Numbers of Clusters.}
We first analyze the impact of the number of clusters $k_{\text{ssl}}$ used in k-means.
The results are shown in the first and second rows of Table~\ref{tbl:analysis-ablation}.
When $k_{\text{ssl}}$ is set to 39 or 100, the performance across the four downstream tasks degrades significantly.
As we have alluded to in the introduction, we attribute this to the problem
of not being able to identify enough neurons to protect.
Figure~\ref{fig:num-ivector-neurons} shows how the number of identified neurons changes
when we use different cluster sizes, and we can see that when the number of clusters
is large, few neurons are identified except for the first and the last layers.

\vspace{0.5em}
\noindent\textbf{Effect of Removing SSL or i-vector Cluster Neurons.}
Next, we evaluate the impact of not protecting either SSL cluster neurons or i-vector cluster neurons.
The results are presented in the third and fourth rows of Table~\ref{tbl:analysis-ablation}.
We observe that both types of neurons are essential for maintaining performance, as the removal of either negatively affects the results.
Interestingly, SSL cluster neurons appear to be more critical than i-vector cluster neurons for SID.
This finding aligns with previous research suggesting that phonetic information is useful for the learning of speaker embeddings~\cite{liu2018speaker, liu2019introducing}.

\vspace{0.5em}
\noindent\textbf{Alternatives to Identifying i-vector Cluster Neurons.}
Finally, we explore the alternatives of identifying i-vector cluster neurons
described in Section~\ref{sec:ivector-neurons}.
The results, presented in the fifth, sixth, and seventh rows of Table~\ref{tbl:analysis-ablation}, show that conditioning and then intersecting outperforms the other alternatives
in three out of four tasks, with IC as the only exception.
Overall, conditioning and then intersecting shows better performance than other alternatives.
We again attribute this to the number of neurons identified.
As shown in the bottom part of Figure~\ref{fig:num-ivector-neurons},
the i-vector cluster neurons identified by conditioning and intersecting
are fewer compared to those from Alternative A and Alternative B,
allowing for more efficient pruning that avoids protecting too many neurons.
In contrast, Alternative C fails to identify enough number of i-vector cluster neurons
in the middle layers.

\section{Conclusion}

In this work, we introduce the concept of SSL cluster neurons and i-vector cluster neurons.  
Our approach better retains performance when these neurons are protected during pruning.  
Notably, it is particularly effective for one-shot pruning, even surpassing the unpruned results in some cases.  
Additionally, we conduct analysis and ablation studies on the neuron identification process, addressing gaps in prior research.  
Overall, these cluster neurons play a significant role in encoding speaker information.  

\printbibliography

\end{document}